\documentclass{article}
\usepackage{ijcai24}

\usepackage{times}
\usepackage{soul}
\usepackage{url}
\usepackage[hidelinks]{hyperref}
\usepackage[utf8]{inputenc}
\usepackage[small]{caption}
\usepackage{graphicx}
\usepackage{amsmath}
\usepackage{amsthm}
\usepackage{booktabs}
\usepackage{multirow}
\usepackage{bm}
\usepackage{bbding}
\usepackage{pifont}
\usepackage{makecell}
\usepackage{xcolor}
\usepackage{array}
\usepackage{tabularx}
\usepackage{algorithm}
\usepackage{algorithmic}
\usepackage[switch]{lineno}
\usepackage{amsmath} % assumes amsmath package installed
\usepackage{amssymb}  % assumes amsmath package installed
\usepackage{booktabs}
\usepackage{multirow}
\usepackage[hidelinks]{hyperref}
\usepackage{pifont}

\title{Characterized Diffusion and Spatial-Temporal Interaction Network for Trajectory Prediction in Autonomous Driving}

\author{
Haicheng Liao\textsuperscript{\rm 1}\thanks{Authors contributed equally; \dag Corresponding author.}\and
Xuelin Li\textsuperscript{\rm 2}$^{*}$\and
Yongkang Li\textsuperscript{\rm 2}\and
Hanlin Kong\textsuperscript{\rm 2}\and
Chengyue Wang\textsuperscript{\rm 1}\and\\
Bonan Wang\textsuperscript{\rm 1}\and
Yanchen Guan\textsuperscript{\rm 1}\and
KaHou Tam\textsuperscript{\rm 1}\and
Zhenning Li\textsuperscript{\rm 1}$^{*\dag}$\and
Chengzhong Xu\textsuperscript{\rm 1}
\\
\affiliations
$^1$University of Macau \\
$^2$University of Electronic Science and Technology of China
\emails
\{yc27979, chengyuewang, mc3500, yc37976, yc374361, zhenningli, czxu\}@um.edu.com,
lxl.cooper@outlook.com,
franklin1234560@163.com,
hanlinkong@foxmail.com
}

\begin{document}

\maketitle

\begin{abstract}
Trajectory prediction is a cornerstone in autonomous driving (AD), playing a critical role in enabling vehicles to navigate safely and efficiently in dynamic environments. To address this task, this paper presents a novel trajectory prediction model tailored for accuracy in the face of heterogeneous and uncertain traffic scenarios. At the heart of this model lies the Characterized Diffusion Module, an innovative module designed to simulate traffic scenarios with inherent uncertainty. This module enriches the predictive process by infusing it with detailed semantic information, thereby enhancing trajectory prediction accuracy. Complementing this, our Spatio-Temporal (ST) Interaction Module captures the nuanced effects of traffic scenarios on vehicle dynamics across both spatial and temporal dimensions with remarkable effectiveness. Demonstrated through exhaustive evaluations, our model sets a new standard in trajectory prediction, achieving state-of-the-art (SOTA) results on the Next Generation Simulation (NGSIM), Highway Drone (HighD), and Macao Connected Autonomous Driving (MoCAD) datasets across both short and extended temporal spans. This performance underscores the model's unparalleled adaptability and efficacy in navigating complex traffic scenarios, including highways, urban streets, and intersections.
\end{abstract}
\begin{figure}[t]
  \centering  \includegraphics[width=1\linewidth]{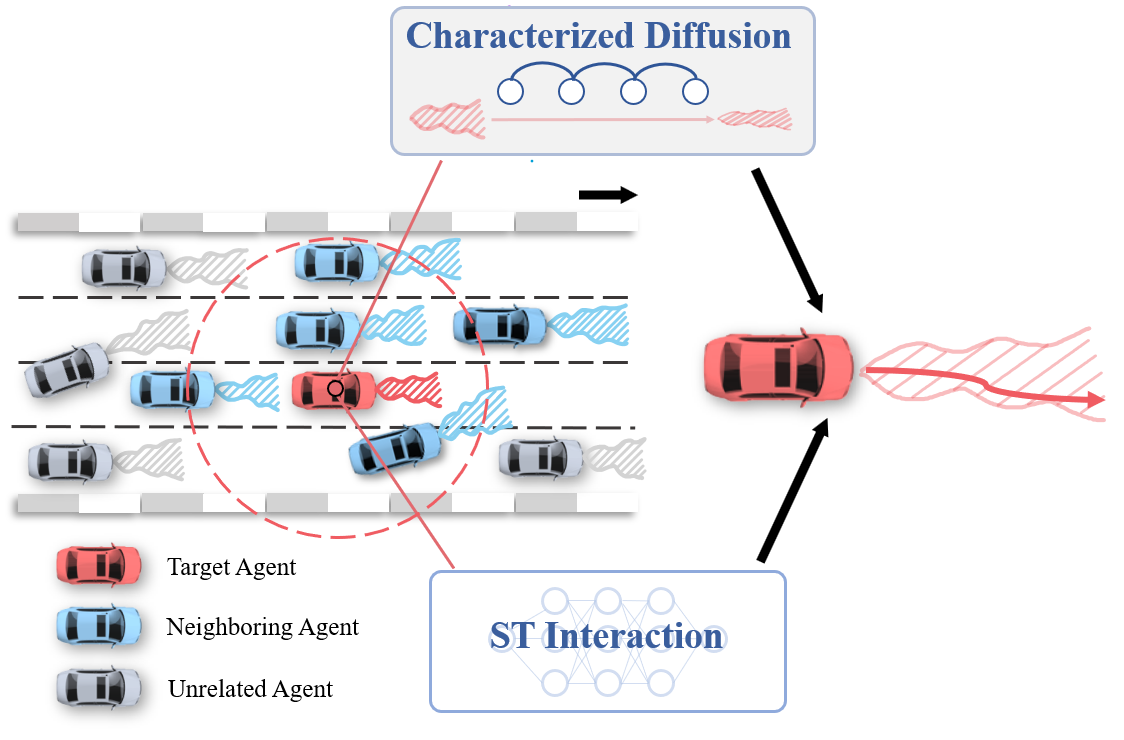} 
  \caption{Overview of our model for processing past states. The framework utilizes two specialized modules to accomplish trajectory prediction for the target agent: Characterized Diffusion and Spatial-Temporal Interaction Network. In situations of high uncertainty, characterized diffusion employs a noisy Gaussian function to define a confidence region for the trajectory distribution. Continuous denoising isolates confidence features for future predictions. Meanwhile, the spatial-temporal interaction network extracts features to understand spatial relations and temporal dependency.}
  \label{head}
\end{figure}

\section{Introduction}

In the domain of autonomous driving (AD), trajectory prediction plays a pivotal role by providing invaluable insights for the subsequent trajectory planning module, thereby enhancing the safety of navigation in complex and dynamic traffic scenarios \cite{huang2022survey}. The continuous presence of mixed traffic flow necessitates a trajectory prediction model that deeply understands the heterogeneity and uncertainty of traffic scenarios \cite{liao2024human}. Despite the proliferation of trajectory prediction models, significant gaps remain in the thorough investigation of the impact of heterogeneous and uncertain traffic scenarios on future motion.

The initial gap we pinpoint hinges on the accurate simulation of future traffic scenarios—a cornerstone for enhancing trajectory prediction precision \cite{wang2023wsip,liao2024cognitive}. The challenge is amplified by the intrinsic uncertainties characterizing traffic dynamics, making the accurate forecast of future scenarios a complex endeavor. Prevailing models have primarily concentrated on uncertainties inherent to the target agent \cite{zhao2019multi,alahi2016social,gupta2018social}, thereby neglecting the comprehensive uncertainty pervasive in the overall traffic scenarios. This oversight highlights an imperative need for trajectory prediction frameworks to adeptly navigate and mitigate uncertainties, facilitating a more accurate and holistic simulation of future traffic scenarios.

The second identified gap related to the sophisticated mechanisms by which traffic scenarios exert influence on human driving behaviors. The decision-making processes of human drivers are profoundly shaped by their interactions with other traffic agents, with such interactions predicated on a nuanced interplay between spatial and temporal dimensions. Nevertheless, prevailing models have primarily focused on capturing spatial interactions, starkly overlooking the critical temporal dynamics \cite{wang2023wsip}. This neglect highlights the critical necessity for models that adeptly incorporate both spatial and temporal interaction. 

To address these challenges, we introduce a novel generative model, CDSTraj, which is built on a dual architecture. As shown in Fig. \ref{head}. Our model employs an encoder designed to generate spatial-temporal features from past states. while significantly fusing confidence features to ensure the stability and reliability of the representation. This enriched feature set serves as the foundational input for a decoder that generates future trajectory predictions. A key innovation in our approach is the introduction of a characterized diffusion, which is seamlessly integrated with a spatial-temporal interaction network. This synergistic combination allows our model to be aware of the indeterminacy associated with both scene-to-agent and agent-to-agent contexts. Consequently, this leads to trajectory predictions that are both more accurate and reliable, even in dynamically changing environments. Overall, our main contributions can be summarized as follows:
\begin{itemize}
    \item  We introduce the Characterized Diffusion Module, a novel approach that enhances trajectory prediction by dynamically simulating future traffic scenarios through iterative uncertainty mitigation. This module significantly augments the predictive accuracy by integrating complex, contextual scenario features, allowing for a more nuanced understanding of potential motion.
    
    \item  We unveil the Spatial-Temporal Interaction Module, which leverages a spatio-temporal attention mechanism to meticulously model and analyze the intricate interactions characteristic of traffic scenarios. Unique to this module is its three-stage architecture, meticulously designed to efficiently capture and process information across spatial and temporal dimensions.
    
    \item  Our rigorous empirical investigations underscore the superiority of our model over the existing trajectory prediction model. Through extensive experiments, our model achieves top performance on public datasets such as NGSIM, HighD, and MoCAD. The exceptional performance on the MoCAD dataset is especially significant, offering a fresh perspective for evaluation through its unique right-hand drive configuration and obligatory left-hand traffic regime, thus underscoring our model's adaptability and accuracy in varying driving scenarios.
\end{itemize}

\section{RELATED WORKS}
\textbf{Trajectory Prediction for Autonomous Driving.}
Early trajectory prediction methods primarily relied on manual feature engineering and rule-based techniques, including linear regression and Kalman filters \cite{prevost2007extended}. These methods were limited in capturing complex interactions in dynamic environments. The field evolved significantly with the introduction of deep learning, specifically Recurrent Neural Networks (RNNs) \cite{kim2017probabilistic} and Long Short-Term Memory (LSTM) networks \cite{altche2017lstm,alahi2016social,liao2024bat}. These advancements enabled the capture of temporal dependencies in trajectories. Further innovation came with Graph Neural Networks (GNNs) \cite{zhou2021ast,10468619,liao2024mftraj}, which provided a more nuanced approach to modeling interactions among agents in crowded scenes.

\textbf{Generative Models for Trajectory Prediction.}
Generative models like Generative Adversarial Networks (GANs) \cite{gupta2018social} and Variational Auto-Encoders (VAEs) \cite{lee2017desire} have gained prominence in trajectory prediction. GANs involve a generator and a discriminator engaged in mutual learning, while VAEs use a generative model and a variational posterior, the optimization of which can be complex. Diffusion models, on the other hand, offer a simplified training process by focusing on matching forward and inverse diffusion processes. To the best of our knowledge, this work is the first to leverage diffusion models for capturing the confidence features.
\begin{figure*}[t]
  \centering
\includegraphics[width=0.9\linewidth]{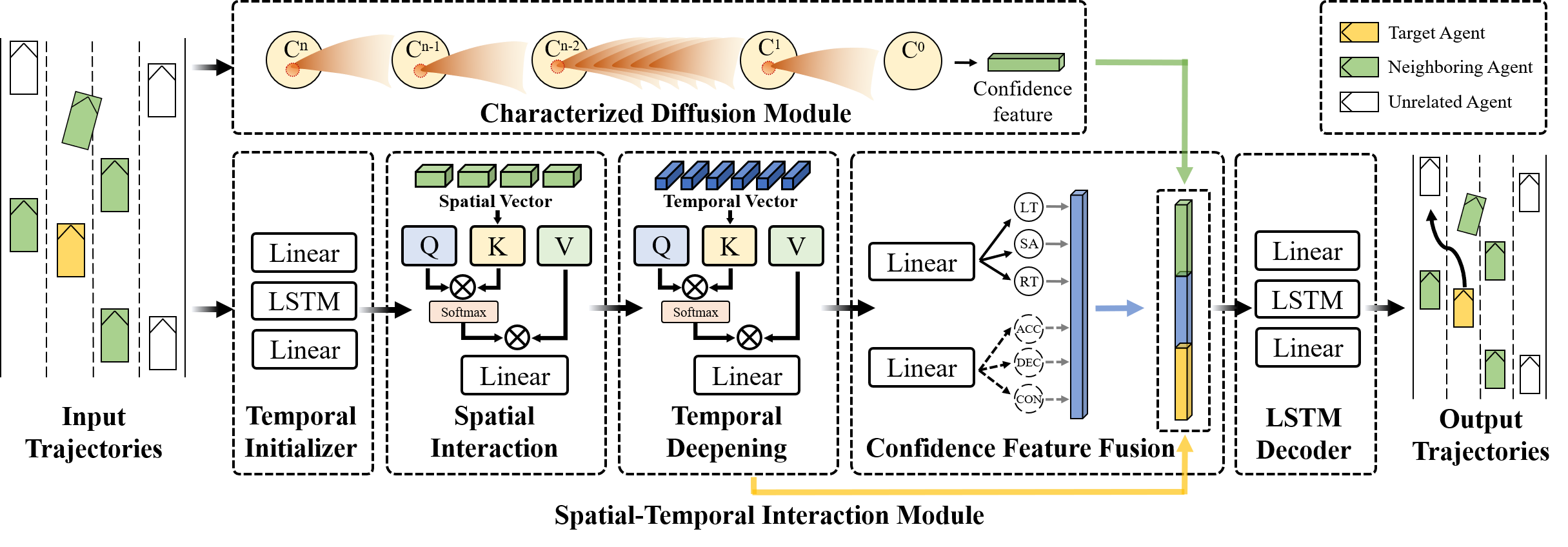}
  \caption{Framework of the proposed model.}
  \label{model}
\end{figure*}

\textbf{Denoising Diffusion Probabilistic Models.}
Denoising Diffusion Probabilistic Models (DDPM) \cite{ho2020denoising}, known as diffusion models, have gained prominence as powerful generative models for various applications, including image \cite{ramesh2022hierarchical,rombach2022high,liao2024gpt}, video \cite{ho2022imagen}, and 3D shape\cite{poole2022dreamfusion} generation. Inspired by the diffusion models' enormous representation capacities in numerous generation tasks, our work introduces the novel application of diffusion models to trajectory prediction in autonomous driving, addressing the challenges of modeling uncertainties and complex agent interactions in dynamic environments.  
\section{Problem Formulation}\label{Problem}
The paramount objective of this study is the precise prediction of trajectories for all entities within the proximity of an autonomous vehicle (AV) situated in an environment characterized by mixed autonomy. For this purpose, every entity proximal to the AV is designated as a \textit{target agent}. At a specific time $t_c$, our model endeavors to utilize the historical states of both the target agent and its neighboring agents to predict the future trajectory of the target agent, represented as $\bm{Y}_{0}$, extending to a future time $t_{c}+t_{f}$. The historical state since the time $t_{c}-t_h$ is denoted by $\bm{X}_{0}$ for the target agent and $\bm{X}_{i}$ for the neighboring agents.

The novelty of our model lies in its exploitation of anticipated future traffic scenarios, specifically the future trajectories of neighboring agents, to enhance the accuracy of trajectory prediction for the target agent. To this end, we develop an Characterized Diffusion Module, designed to systematically mitigate the uncertainty inherent in the trajectories of neighboring agents, thereby enabling the accurate prediction of their future trajectories $\bm{Y}_{i}$. Formally, our prediction model $\Phi$ is represented as:
\begin{equation}\label{eq.1}
    \bm{Y}_{0} = \Phi(\bm{X_0}, \bm{X_i}, \bm{Y_i})\  \; \forall i\in[1,n] 
\end{equation}

\section{Methodology}\label{Methodology}

\subsection{Overview of the Model Framework} 
As illustrated in Figure \ref{model}, our model comprises three primary components: the Characterized Diffusion Module, the ST Interaction Module, and a Multi-modal Decoder. The Characterized Diffusion Module employs an inverse diffusion process to generate the future trajectories of neighboring agents. Concurrently, the ST Interaction Module extracts Spatial-Temporal Interaction features through a methodical alternation between spatial and temporal dimensions. Ultimately, the predicted trajectories are generated using a multi-modal decoder, which synthesizes the processed information to produce accurate trajectory prediction.

\subsection{Characterized Diffusion} 
\begin{figure*}[t]
      \centering
    \includegraphics[width=0.8\linewidth]{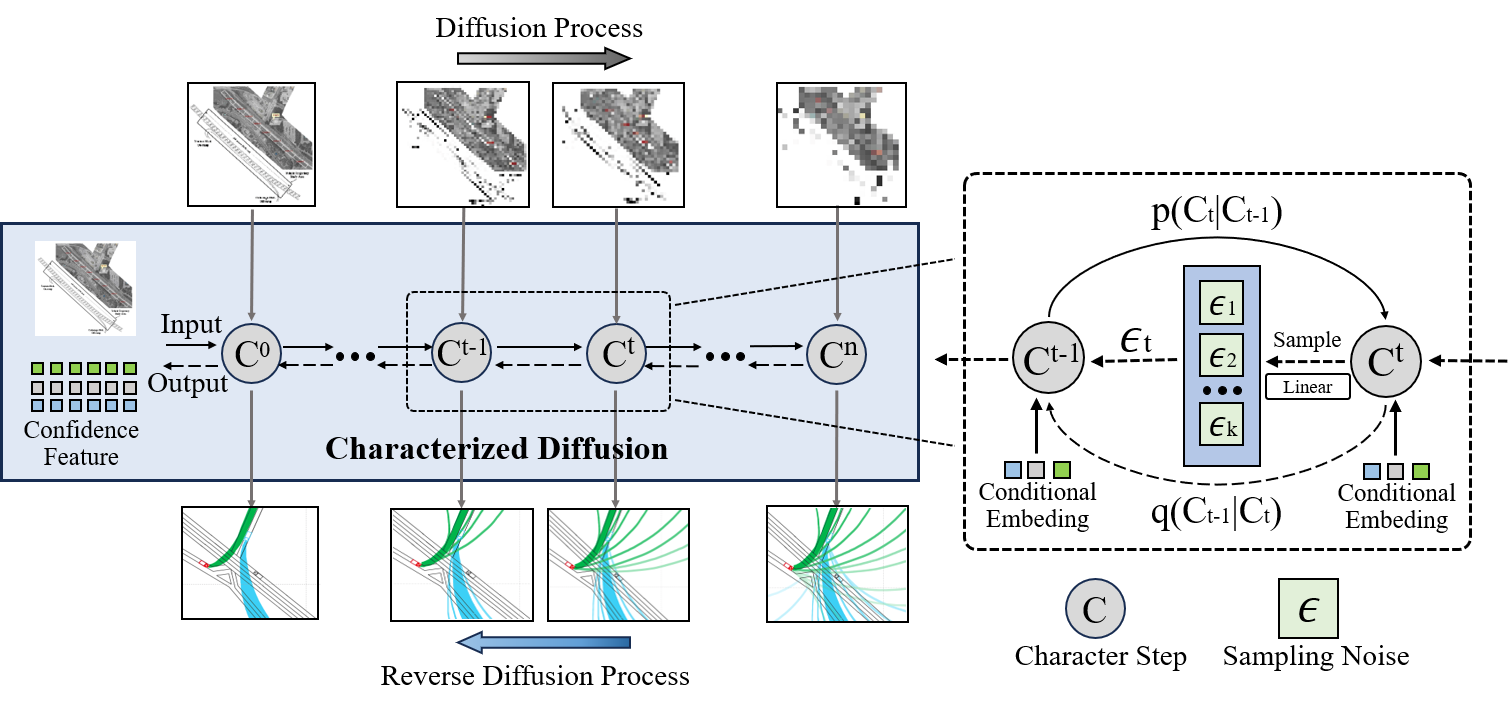}
      \caption{Overview of the Characterized Diffusion Module. The observed historical states are used as input for forward diffusion, noise is added for every steps by the Gaussian distributions with controllable means and variances. When $t = n$, the model performs reverse diffusion guided by the conditional embedding between each step to estimate the noise from the sampled noises with a Linear network, and finally output future trajectories.}
      \label{figurelabel2}
\end{figure*}

In order to predict the future trajectories of neighboring agents, the Characterized Diffusion Module treats trajectory prediction as the reverse process of motion Characterized Diffusion and gradually eliminates the uncertainty of future trajectories by learning a parameterized Markov chain with the observed historical states. More specifically, during the diffusion process, the uncertainty inherent in future trajectories is simulated by iteratively introducing Gaussian noise. Conversely, in the inverse diffusion process, this uncertainty is iteratively mitigated to derive the anticipated future trajectories accurately. The detailed procedure is shown in Fig. \ref{figurelabel2}. Mathematically, let $\textbf{C}$ be the future trajectory of the neighboring agents. Firstly, We initialize the diffused unit $\textbf{C}^0$:
\begin{equation}
\textbf{C}^0 = \textbf{C}
\end{equation}

We use a forward diffusion operation $f_\textit{diffuse}(\cdot)$ to add uncertainty to $\textbf{C}^{\delta-1}$ and transition to diffused unit $\textbf{C}^\delta$:
\begin{equation}
\textbf{C}^\delta = f_\textit{diffuse}(\textbf{C}^{\delta-1}), \delta = 1,...,\Gamma
\end{equation}
where $\textbf{C}^{\delta}$ is the diffused unit at the $\delta^{th}$ diffusion step, $\Gamma$ is the total number of steps. 

After $n$ iterations, our model is able to capture a comprehensive spectrum of uncertain traffic scenarios with maximum coverage. Then, the inverse diffusion process is applied to accurately derive the future trajectories of neighboring vehicles. This step-by-step refinement ensures high fidelity in predicting vehicle movements, effectively addressing the inherent complexity of dynamic traffic scenarios. Due to the indeterminacy of future trajectories, in the denoising procedure, it is usually more reliable to capture more than one inverse unit to extract sufficient trajectory information. Therefore, we draw $K$ independent and identically distributed samples to initialize the denoising unit $\widehat{\mathbf{C}}_{k}^{\Gamma}$ from a normal distribution.
\begin{equation}
\widehat{\mathbf{C}}_{k}^{\Gamma} \stackrel{i . i . d}{\sim} \mathcal{P}\left(\widehat{\mathbf{C}}^{\Gamma}\right)=\mathcal{N}\left(\widehat{\mathbf{C}}^{\Gamma}; \mathbf{0},\mathbf{I}\right), \text { sample } K \text { times, } 
\end{equation}

We formulate the trajectories generation process as a reverse diffusion by iteratively applying a denoising operation $f_\textit{denoise}(\cdot)$ to obtain the denoised unit $  \widehat{\mathbf{C}}_{k}^{\delta}$ conditioned on historical states $\textbf{X}_0$, $\textbf{X}_{i}$ and, unit $ \widehat{\mathbf{C}}_{k}^{\delta+1}$:
\begin{equation}
\widehat{\mathbf{C}}_{k}^{\delta}=f_{\textit{denoise }}\left(\widehat{\mathbf{C}}_{k}^{\delta+1}, \mathbf{X}_0, \textbf{X}_{i}\right), \delta=\Gamma-1, \cdots, 0,
\end{equation}

In a denoising module, two parts are trainable: a transformer-based context encoder $f_\textit{context}(\cdot)$ to learn a social-temporal embedding and an uncertainty estimation module $f_\epsilon(\cdot)$ to estimate the uncertainty to reduce. Mathematically, the $\delta^{th}$ denoising step works as follows:
\begin{equation}
 \mathbf{C}_\textit{encoder}=f_{\textit{context}}\left(\mathbf{X}_0, \mathbb{X}_{i}\right)
\end{equation}
\begin{equation}\label{eq7}
\boldsymbol{\epsilon}_{\theta}^{\delta}=f_{\boldsymbol{\epsilon}}\left(\widehat{\mathbf{C}}_{k}^{\delta+1}, \mathbf{C}_\textit{encoder}, \delta+1\right)
\end{equation}
\begin{equation}
\widehat{\mathbf{C}}_{k}^{\delta}=\frac{1}{\sqrt{\alpha_{\delta}}}\left(\widehat{\mathbf{C}}_{k}^{\delta+1}-\frac{1-\alpha_{\delta}}{\sqrt{1-\bar{\alpha}_{\delta}}} \boldsymbol{\epsilon}_{\theta}^{\delta}\right)+\sqrt{1-\alpha_{\delta}} \mathbf{z}
\end{equation}
where $\alpha_{\delta}$ and $\bar{\alpha}_{\delta}=\prod_{i=1}^{\delta} \alpha_{i}$ are parameters in the diffusion process and $\mathbf{z} \sim \mathcal{N}(\mathbf{z} ; \mathbf{0}, \mathbf{I})$ is the uncertainty. We leverage a context encoder $f_{\text {context }}(\cdot)$ on historical states $\left(\mathbf{X}, \mathbb{X}_{\mathcal{N}}\right)$ to obtain the context condition $\mathbf{C}_{encoder}$, and estimates the uncertainty $\epsilon_{\theta}^{\delta}$ in the uncertain trajectory $\widehat{\mathbf{Y}}_{k}^{\delta+1}$ through uncertainty estimation $f_{\epsilon}(\cdot)$ implemented by multi-layer perceptions with the context C; Eq. \ref{eq7} provides a standard denoising step.

The final ${K}$ extracted units are $  \widehat{\mathbf{C}} = \{\widehat{\mathbf{C}}_{1}^{0},\widehat{\mathbf{C}}_{2}^{0},...,\widehat{\mathbf{C}}_{K}^{0}\}.$ Finally, we obtain the future trajectories with:
\begin{equation}
\textbf{Y}_i = \Omega(\widehat{\mathbf{C}}, W_{cf})
\end{equation}
where $\Omega$ denotes a multi-layer perception with learnable parameter matrix $W_\textit{cf}$.

\subsection{Spatial and Temporal Interaction}
To enhance the precision of modeling the temporal and spatial dynamics of vehicle interactions within the environment, the Space-Time (ST) Interaction Module is designed with a novel structure that alternates between temporal and spatial dimensions. This module is composed of three key components, as illustrated in Fig. \ref{model}:  1. \textbf{Temporal Encoder:} where the temporal dependencies of all the agents are extracted from the historical states by a temporal encoder. 2. \textbf{Spatial Encoder:}, which plays an essential role in extracting the spatial relation between the target agent and neighboring agents. 3. \textbf{ST Fusion:} which aims to deeply capture the spatial-temporal interaction.

\indent 1) $\textbf{Temporal  Encoder:}$ To begin with, a temporal embedding vector $F^t$ is obtained from the historical states $x^t$ at the $t^{th}$ timestamp using a fully connected layer with a learnable parameter matrix $W_\textit{emb}$ as follows:
\begin{equation}
F^t = \delta(\phi(x^t, W_{emb}))
\end{equation}
where $\delta$ is the fully connected layer and $\Phi$ is the LeakyReLU activation function. It can be defined as follows:
\begin{equation}
h^t = f_\textit{tem}(F^t, h^{t-1},W_{init})
\end{equation}
where $W_{init}$ denotes the learnable parameter matrix of encoder $f_\textit{tem}$ and $h^t$ denotes the temporal feature at the $t^{th}$ timestamp, which is updated at each timestamp based on the hidden state at the previous timestamp and the embedding vector at the current timestamp. Here, we apply $f_\textit{tem}$ to every single agent shared with the parameters of encoding $f_\textit{tem}$ to reduce variation across numerous agents at different timestamps. Finally, for the target agent, we obtain $H_0 = [h^{-T_p+1}_0,h^{-T_p+2}_0,...,h^0_0] \in R^{T_p \times D}$ representing the temporal feature over $T_p$ timestamps, and $\bar{H}_i = [\bar{h}^{-T_p+1}_i,\bar{h}^{-T_p+2}_i,...,\bar{h}^0_i] \in R^{T_p \times D}$ denoting the temporal feature of $i^{th}$ neighboring agents and $D$ denotes the number of hidden dimensions.

\indent 2) $\textbf{Spatial Encoder}$: Apparently, naive use of one temporal feature per agent does not capture agent-to-agent spatial relations. Therefore, capturing the spatial relations between agents in the same scene is highly necessary. Given the success of attention mechanisms in sequence-based prediction \cite{hu2020introductory}, we adopt a multi-head attention mechanism to obtain spatial relations between agents, which can be represented as follows:
\begin{equation}
Q,K,V=f_{sp}(H,\bar{H},W_{q},W_{k},W_{v})
\end{equation}
where $f_{sp}$ is the spatial attention.

In a detailed elaboration, $Q=[{q^{-T_p+1},q^{-T_p+2},...,q^0}]$, $K=[k^{-T_p+1},k^{-T_p+2},...,k^0]$, $V=[v^{-T_p+1},v^{-T_p+2},...,v^0]$ respectively denote linear projected vectors query, key, value and $W_q$,$W_k$,$W_v$ indicate three learnable parameter matrices. We apply normalization $\Pi$ to query and key to represent the importance of agents influencing each other. Formally,
\begin{equation}
\omega= \Pi(\frac{(Q\cdot K)}{\sqrt{D_{init}}})
\end{equation}
where $\omega$ is the attention score representing the similarity between $Q$ and $K$, 
 $(\cdot)$ denotes the operation of matrix multiplication. Subsequently, we leverage attention scores to discern the significant connections among agents. Mathematically,
\begin{equation}
\Upsilon = \omega \cdot V
\end{equation}
where $\upsilon$ is the output of the single-head attention. Compared to single-head, multi-head attention can comprehensively capture local and global spatial relations. Therefore, we apply the multi-head attention mechanism to obtain $\Upsilon=[\upsilon_1,\upsilon_2,...,\upsilon_n]$, carrying spatial relations. We introduce an innovative gating mechanism $H_g$ to control the importance of different heads and selectively amplify or suppress specific heads. This mechanism acts as a gatekeeper for the $\Upsilon$ by adjusting the activation level consisting of two linear layers as:
\begin{equation}
H_a = \kappa (\Upsilon), \,\,
H_g = \sigma(\kappa(\Upsilon)), \,\,
S = H_a \odot H_g 
\end{equation}
where $\sigma$ denotes the activation function sigmoid and $\odot$ denotes the multiplication of the corresponding elements of the matrix, $\kappa$ denotes the linear layer. Note that the output of the spatial encoder we simplify as follows:
\begin{equation}
S = [s^{-T_p+1},s^{-T_p+2},...,s^{0}] \in R^{T_p \times D}
\end{equation}

\indent 3) $\textbf{ST Fusion}$: Following the spatial encoder, we introduce a ST Fusion $f_\textit{ST}$ to deeply capture spatial-interaction from $S$ produced by the preceding module. Formally,
\begin{equation}
\bar{Q},\bar{K},\bar{V} = f_\textit{ST}(S,\bar{W}_q,\bar{W}_k,\bar{W}_v) 
\end{equation}
where $\bar{Q} = [\bar{q}^t_{1},\bar{q}^t_{2},...,\bar{q}^t_{M}]$, $\bar{K}=[\bar{k}^t_1,\bar{k}^t_2,...,\bar{k}^t_M]$, $\bar{V} = [\bar{v}^t_1,\bar{v}^t_2,...,\bar{v}^t_M]$ respectively denote linear projected vectors query, key, value at the $t^{th}$ timestamp, and $\bar{W}_q$,$\bar{W}_k$,$\bar{W}_v$ indicate three learnable parameter matrices. In analogy with the spatial interaction module, we also use normalization and gating mechanisms to obtain long-term temporal features simplified as follows:
\begin{equation}
U = [u^{-T_p+1}, u^{-T_p+2},...,u^0] \in R^{T_p \times D}
\end{equation}

\subsection{Decoder}
This study defines the trajectory prediction task as a conditional probabilistic prediction problem. Specifically, the decoder is designed to predict the future trajectory for the target agent based on different lateral and longitudinal maneuver classes:
\begin{equation}
P(\hat{Y}) = P(Y|P_{lat},P_{lon})\cdot P_{lat} \cdot P_{lon}
\end{equation}
where $P({\hat{Y}})$ is the conditional probabilistic distribution for the predicted trajectory. In detail, we use the LSTM decoder to implement the final multi-modal trajectory prediction.
\begin{equation}
\hat{y^t} = f_\textit{LSTM}(F,\hat{y}^{t-1},W_\textit{decoder})
\end{equation}
where $\hat{y}^t$ is the predicted 2D spatial coordinate at future $t^{th}$ timestamp, $W_{decoder}$ denotes the parameter matrix to be learned in the LSTM. Based on the comprehensive feature vectors, we can only apply the simplest LSTM decoder to predict the accurate trajectory easily with fewer parameters.

\section{Experiment}\label{Experiments}
To evaluate the performance of our model, we perform expensive experiments on real-world datasets. This study uses a consistent segmentation framework for all three datasets. Each sample is divided into 8-second segments, with the first 16 timestamps (3 seconds) serving as historical data and the following 25 timestamps (5 seconds) for evaluation. 

\subsection{Datasets}
\textbf{Next Generation Simulation (NGSIM):} This dataset \cite{deo2018multi} consists of vehicle trajectory datasets from US-101 and I-80, containing approximately 45 minutes of vehicle trajectory data at 10 Hz. It is critical for the analysis of vehicle behavior in a variety of traffic scenarios and assists in the development of reliable AD models.

 \textbf{Highway Drone (HighD):} HighD \cite{krajewski2018highd} is a dataset of vehicle trajectories collected from six locations on German highways. It includes 110,000 vehicles, including cars and trucks, and a total distance traveled of 45,000 km. This dataset provides detailed information about each vehicle, including type, size, and maneuvers, making it invaluable for advanced vehicle trajectory analysis and AD research.

\textbf{Macau Connected Autonomous Driving (MoCAD):} It \cite{liao2024bat} was collected from the first Level 5 autonomous bus in Macau, which has undergone extensive testing and data collection since its deployment in 2020. The data collection period spans over 300 hours and covers various scenarios, including a 5-kilometer campus road dataset, a 25-kilometer dataset covering city and urban roads, and complex open traffic environments captured under different weather conditions, time periods, and traffic densities.

\begin{table}[htbp]
  \centering
  \caption{Evaluation of the proposed model and baselines on the NGSIM dataset over a 5-second prediction horizon. The accuracy metric is RMSE (m). Cases marked as ('-') indicate unspecified values. \textbf{Bold} and \underline{underlined} values represent the best and second-best performance in each category.}
  \resizebox{\linewidth}{!}{
\setlength{\tabcolsep}{3mm}
    \begin{tabular}{cccccc}
 \toprule
    \multirow{2}[3]{*}{Model} & \multicolumn{5}{c}{Prediction Horizon (s)} \\
\cmidrule{2-6}          & 1     & 2     & 3     & 4     & 5 \\
    \midrule
    S-LSTM \cite{alahi2016social}& 0.65  & 1.31  & 2.16  & 3.25  & 4.55  \\
    S-GAN \cite{gupta2018social}& 0.57  & 1.32  & 2.22  & 3.26  & 4.40  \\
    CS-LSTM \cite{deo2018convolutional}& 0.61  & 1.27  & 2.09  & 3.10  & 4.37  \\
    MATF-GAN \cite{zhao2019multi}& 0.66  & 1.34  & 2.08  & 2.97  & 4.13  \\
    DRBP\cite{gao2023dual}& 1.18  & 2.83  & 4.22  & 5.82  & - \\
     M-LSTM \cite{deo2018multi}& 0.58  & 1.26  & 2.12  & 3.24  & 4.66  \\
    IMM-KF \cite{lefkopoulos2020interaction}& 0.58  & 1.36  & 2.28  & 3.37  & 4.55  \\
    GAIL-GRU \cite{kuefler2017imitating}& 0.69  & 1.51  & 2.55  & 3.65  & 4.71  \\
    MFP \cite{tang2019multiple}& 0.54  & 1.16  & 1.89  & 2.75  & 3.78  \\
    NLS-LSTM \cite{messaoud2019non}& 0.56  & 1.22  & 2.02  & 3.03  & 4.30  \\
    MHA-LSTM \cite{messaoud2021attention}& 0.41  & 1.01  & 1.74  & 2.67  & 3.83  \\
    WSiP \cite{wang2023wsip}& 0.56  & 1.23  & 2.05  & 3.08  & 4.34  \\
    CF-LSTM \cite{xie2021congestion}& 0.55  & 1.10  & 1.78  & 2.73  & 3.82  \\
    TS-GAN \cite{wang2022multi}& 0.60  & 1.24  & 1.95  & 2.78  & 3.72  \\
    STDAN \cite{chen2022intention}& 0.42  & 1.01  & 1.69  & 2.56  & 3.67  \\
    BAT \cite{liao2024bat}& \textbf{0.23} & \textbf{0.81}  & \underline{1.54}  & \underline{2.52} &3.62\\
    FHIF \cite{zuo2023trajectory} &0.40  & 0.98  & 1.66  & \underline{2.52}  & 3.63\\ 
    DACR-AMTP \cite{cong2023dacr}& 0.57  & 1.07  & 1.68  & 2.53  & \underline{3.40} \\ 
    \midrule
    \textbf{Our model} & \underline{0.36} & \underline{0.86} & \textbf{1.36} & \textbf{2.02} & \textbf{2.85}  \\
    \bottomrule
    \end{tabular}%
    }
  \label{table_3}%
\end{table}%

\subsection{Training and Implement Details}
We adopt a two-stage training approach to train our model. In the first stage, our model is trained to predict a future trajectory with the Mean Squared Error (\textit{MSE}) \cite{pan2020multiple} loss function as follows:
\begin{equation}
\mathcal{L}_\textit{MSE}(\hat{y}, y) = \sum_{t=1}^{t_f}[(\hat{y^t_x} - y_x^t)^2 + (\hat{y_y^t} - y_y^t)^2] 
\end{equation}
where $(\hat{y}_x^t, \hat{y}_y^t)$ is the 2D spatial coordinate predicted and $(y_x^t,y_y^t)$ denotes the corresponding ground-truth coordinate. It helps our model to learn the exact position information that approximates the true trajectory. When the model converges to a point using the \textit{MSE} loss function, we use a Negative Log-Likelihood (\textit{NLL}) \cite{kim2020multi} loss function. This transition facilitates a more comprehensive exploration of uncertainty within the trajectory prediction.
\begin{align}
Loss_{NLL}(\hat{y}, y) = & \sum_{t=1}^{T_F}\alpha((\sigma^t_x)^2  (\Delta_x^t)^2 +(\sigma^t_y)^2(\Delta_y^t)^2 \nonumber \\
& -2\rho_{xy}^t\sigma^t_x\sigma^t_y (\Delta_x^t)(\Delta_y^t))-log(P^t) \nonumber 
\end{align}
where $\Delta_x^t$,$\Delta_y^t$ represents  $(y_x^t - \hat{y}_x^t)$, $(y_y^t - \hat{y}_y^t)$, and $\sigma_x^t$ , $\sigma_y^t$ denote the standard deviation of x-coordinate and y-coordinate at the $t$th timestamp. $\rho_{xy}^t$ is the correlation coefficient between the $x$ and $y$ coordinates at $t$th timestamp. $P^t = \rho^T_x \rho^t_y  \sqrt{1 - (\rho_{xy}^t)^{2}}$ is the standard deviation of the probability density function in $x$ and $y$ coordinate. $\alpha$ is an empirical constant value to simplify the calculation.

\begin{table}[t]
  \centering
  \caption{Evaluation of our model and SOTA baselines on HighD.}
\setlength{\tabcolsep}{3mm}
   \resizebox{\linewidth}{!}{
    \begin{tabular}{cccccc}
    \toprule
    \multirow{2}[3]{*}{Model} & \multicolumn{5}{c}{Prediction Horizon (s)} \\
\cmidrule{2-6}          & 1     & 2     & 3     & 4     & 5 \\
    \midrule
    S-LSTM \cite{alahi2016social}& 0.22  & 0.62  & 1.27  & 2.15  & 3.41  \\
    S-GAN \cite{gupta2018social}& 0.30  & 0.78  & 1.46  & 2.34  & 3.41  \\
    WSiP \cite{wang2023wsip}& 0.20  & 0.60  & 1.21  & 2.07  & 3.14  \\
    CS-LSTM \cite{deo2018convolutional}& 0.22  & 0.61  & 1.24  & 2.10  & 3.27  \\
    MHA-LSTM \cite{messaoud2021attention}& 0.19  & 0.55  & 1.10  & 1.84  & 2.78  \\
    NLS-LSTM \cite{messaoud2019non}& 0.20  & 0.57  & 1.14  & 1.90  & 2.91  \\
    DRBP\cite{gao2023dual}& 0.41  & 0.79  & 1.11  & 1.40  & - \\
    EA-Net \cite{cai2021environment} & 0.15  & 0.26  & 0.43  & 0.78  & 1.32  \\
    CF-LSTM \cite{xie2021congestion}& 0.18  & 0.42  & 1.07  & 1.72  & 2.44  \\
    STDAN \cite{chen2022intention}& 0.19  & 0.27  & 0.48  & 0.91  & 1.66  \\
      GaVa \cite{liao2024human}& \underline{0.17}  & \underline{0.24}  & \underline{0.42}  & \underline{0.86}  & \underline{1.31}  \\ 
    \midrule
   \textbf{Our model} & \textbf{0.13}& \textbf{0.21} & \textbf{0.32} & \textbf{0.38} & \underline{1.05} \\
    \bottomrule
    \end{tabular}%
    }
  \label{table_4}%
\end{table}%

\begin{table}[t]
  \centering
  \caption{Evaluation of our model and SOTA baselines on MoCAD.}
  \setlength{\tabcolsep}{3mm}
  \resizebox{\linewidth}{!}{
    \begin{tabular}{cccccc}
    \toprule
    \multirow{2}[3]{*}{Model} & \multicolumn{5}{c}{Prediction Horizon (s)} \\
\cmidrule{2-6}          & 1     & 2     & 3     & 4     & 5 \\
    \midrule
    S-LSTM \cite{alahi2016social} & 1.73  & 2.46  & 3.39  & 4.01  & 4.93 \\
    S-GAN \cite{gupta2018social} & 1.69  & 2.25  & 3.30  & 3.89  & 4.69  \\
    CS-LSTM \cite{deo2018convolutional} & 1.45  & 1.98  & 2.94  & 3.56  & 4.49  \\
    MHA-LSTM \cite{messaoud2021attention} & 1.25  & 1.48  & 2.57  & 3.22  & 4.20  \\
    NLS-LSTM \cite{messaoud2019non} & 0.96  & 1.27  & 2.08  & 2.86  & 3.93\\
    WSiP \cite{wang2023wsip} & 0.70  & 0.87  & 1.70  & 2.56  & 3.47  \\
    CF-LSTM \cite{xie2021congestion} & 0.72  & 0.91  & 1.73  & 2.59  & 3.44 \\
    STDAN \cite{chen2022intention} & 0.62  & 0.85  & 1.62  & 2.51  & 3.32  \\
    HLTP \cite{10468619}  &\underline{0.55} &\textbf{0.76} & \underline{1.44} & \underline{2.39} & \underline{3.21} \\
    \midrule
 \textbf{Our model} & 
\textbf{0.39}& \underline{0.82} & \textbf{1.43} & \textbf{2.08} & \textbf{2.74}\\
    \bottomrule
    \end{tabular}%
    }
  \label{table_6}%
\end{table}%

\subsection{Comparison to State-of-the-arts}
Our model's performance is evaluated against more than 15 state-of-the-art (SOTA) methods on each referenced dataset. 
 The experimental results, presented in Table \ref{table_3}, show that our model provides significant improvements in trajectory prediction over the prevailing SOTA baselines. Using Root Mean Square Error (RMSE) as the evaluation metric, our model consistently outperforms most baselines, achieving improvements of 29\% and 22\% over WSiP and STDAN, respectively, over a 5-second horizon. On the HighD dataset, our model consistently outperforms current SOTA baselines, with average improvements ranging from 43\%-70\% for short-term forecasts (1-3 seconds) and 62\%-78\% for long-term forecasts (4-5 seconds). These improvements highlight the importance of integrating spatio-temporal and confidence features. It is noteworthy that the prediction error of the HighD dataset is significantly lower than that of the NGSIM dataset for all algorithms, probably due to the fact that the HighD dataset provides more accurate trajectory data, including detailed information on location and speed. In addition, the HighD dataset contains approximately twelve times more samples than the NGSIM dataset. In addition, on the MoCAD dataset, our model excels on busy urban roads, outperforming SOTA baselines by at least 37\% for short-term predictions and reducing long-term prediction errors by at least 0.58 metres. These improvements highlight the importance of incorporating characterised diffusion and spatio-temporal interaction networks. In conclusion, our results confirm the effectiveness and efficiency of our model in predicting AV trajectories.

\subsection{Ablation Study}
Table \ref{component} analyzes five critical components: characterized diffusion, temporal initializer, spatial interaction, temporal deepening, and confidence feature fusion modules. It shows that we test six models, labeled Model A through Model F. Evaluations against the NGSIM datasets reveal that the stripped-down versions (Models A-E) consistently underperform compared to Model F, which includes all components.
Importantly, the integration of the characterized diffusion and confidence feature fusion modules significantly enhances performance, underscoring their vital role in improving prediction accuracy. The inclusion of the confidence feature fusion module, which merges spatial-temporal confidence features with maneuver states, enables more precise predictions of the target agent's trajectory, particularly through the incorporation of characterized diffusion.
\begin{table}[htbp]
\caption{Ablation studies for core components in NGSIM dataset. }
\label{component}
\begin{center}
 \setlength{\tabcolsep}{2mm}
 \resizebox{0.95\linewidth}{!}{
\begin{tabular}{ccccccc}
\toprule
\multirow{2}*{Components} & \multicolumn{6}{c}{Ablation Models} \\
\cmidrule{2-7}
& A & B & C & D & E & F\\
\midrule
Characterized Diffusion
& \ding{55} & \ding{52} & \ding{52} & \ding{52} & \ding{52} & \ding{52}\\
Temporal Encoder
& \ding{52} &  \ding{55}  & \ding{52} & \ding{52} & \ding{52} & \ding{52}\\
Spatial Encoder
& \ding{52} & \ding{52} &  \ding{55}  & \ding{52} & \ding{52} & \ding{52}\\
ST Fusion
& \ding{52} & \ding{52} & \ding{52} &  \ding{55}  & \ding{52} & \ding{52}\\
Decoder 
& \ding{52} & \ding{52} & \ding{52} & \ding{52} &  \ding{55} & \ding{52} \\
\hline
RMSE  & 3.09 & 3.05 & 2.97 & 3.02 & 3.16 &2.85\\
\bottomrule
\end{tabular}
}
\end{center}
\end{table}
\begin{figure}[thpb]
      \centering
      \includegraphics[width=0.95\linewidth]{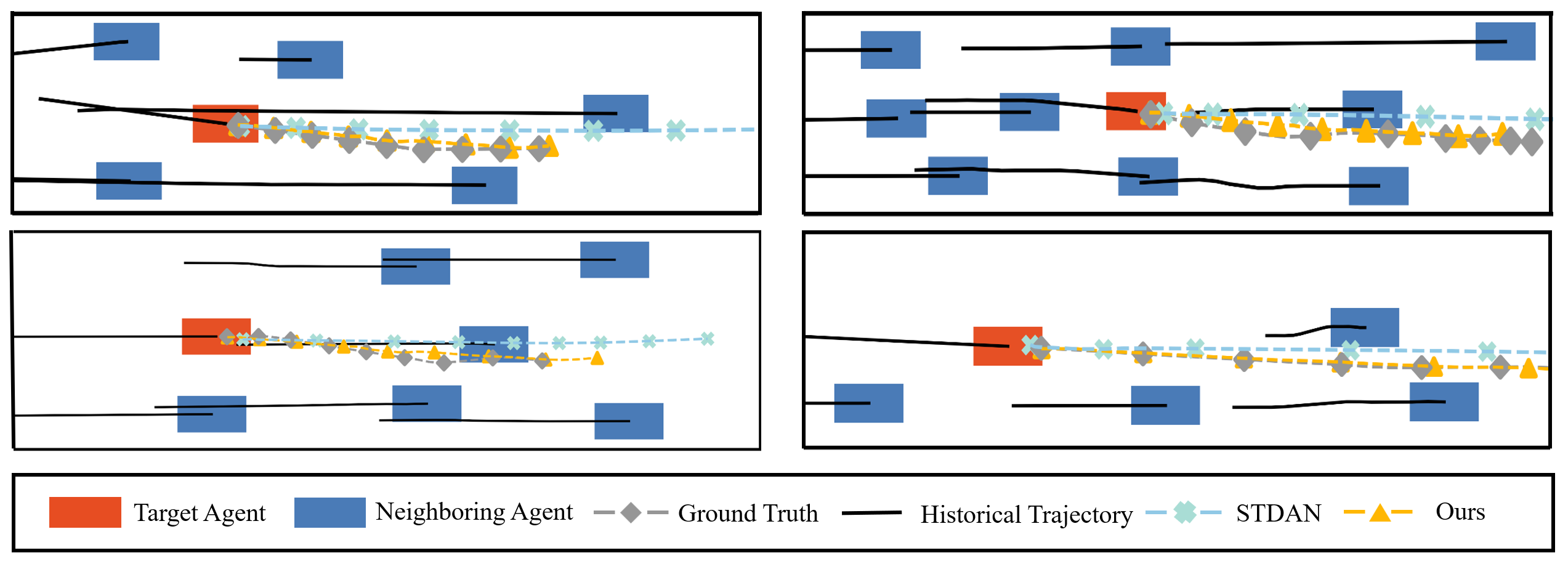}
      \caption{Visualization of our proposed model.}
      \label{figurelabel1}
   \end{figure}
\subsection{Qualitative Results}
To qualitatively gain insight, we visualize the prediction results to assess the effectiveness of our proposed model. Fig. \ref{figurelabel1}. displays the qualitative results of our model tested on the NGSIM dataset. As we can see, the prediction favorably covers the ground truth trajectory despite different numbers of neighboring agents in different scenes. These results demonstrate the impressive capabilities of the CDSTraj model in feature representation.

\section{Conclusion}
The intricate dynamics and uncertainties inherent in multi-agent interactions and scene contexts present a significant challenge in the development of fully AVs. To address this challenge, we introduce a novel generative model that employs a dual architecture integrating a characterized diffusion mechanism and a spatial-temporal interaction network. Empirical evaluations on the NGSIM, HighD, and MoCAD datasets demonstrate that our model, CDSTraj, consistently outperforms existing SOTA baselines in prediction accuracy over both short and long terms. Future research will explore the application of this model to pedestrian trajectory predictions and investigate the integration of spatial and temporal information. These endeavours may yield significant advancements in AD technologies.

\section*{Acknowledgements}
This research is supported by the Science and Technology Development Fund of Macau SAR (File no. 0021/2022/ITP, 0081/2022/A2, 001/2024/SKL), and University of Macau (SRG2023-00037-IOTSC).

\bibliographystyle{named}
\bibliography{ijcai24}

\end{document}